\documentclass[sigconf, authorversion]{acmart}

\begin{document}

\copyrightyear{2025}
\acmYear{2025}
\acmConference[Preprint]{}{September 2025}{}
\acmBooktitle{Preprint} \acmDOI{10.1145/3717511.3749286}
\acmISBN{979-8-4007-1508-2/2025/09}

\title{A Risk Ontology for Evaluating AI-Powered Psychotherapy Virtual Agents}

\author{Ian Steenstra}
\affiliation{%
  \institution{Northeastern University}
  \city{Boston}
  \state{MA}
  \country{USA}
}
\email{steenstra.i@northeastern.edu}

\author{Timothy Bickmore}
\affiliation{%
  \institution{Northeastern University}
  \city{Boston}
  \state{MA}
  \country{USA}
}
\email{t.bickmore@northeastern.edu}

\begin{abstract}
The proliferation of Large Language Models (LLMs) and Intelligent
Virtual Agents acting as psychotherapists presents significant opportunities for expanding mental healthcare access. However, their
deployment has also been linked to serious adverse outcomes, including user harm and suicide, facilitated by a lack of standardized
evaluation methodologies capable of capturing the nuanced risks of
therapeutic interaction. Current evaluation techniques lack the sensitivity to detect subtle changes in patient cognition and behavior
during therapy sessions that may lead to subsequent decompensation. We introduce a novel risk ontology specifically designed for
the systematic evaluation of conversational AI psychotherapists.
Developed through an iterative process including review of the
psychotherapy risk literature, qualitative interviews with clinical
and legal experts, and alignment with established clinical criteria
(e.g., DSM-5) and existing assessment tools (e.g., NEQ, UE-ATR), the
ontology aims to provide a structured approach to identifying and
assessing user/patient harms. We provide a high-level overview
of this ontology, detailing its grounding, and discuss potential
use cases. We discuss four use cases in detail: monitoring real user interactions, evaluation with simulated patients, benchmarking and comparative analysis, and identifying unexpected outcomes. The proposed ontology offers a foundational step towards
establishing safer and more responsible innovation in the domain
of AI-driven mental health support.
\end{abstract}

\ccsdesc[500]{Applied computing~Consumer health}
\ccsdesc[500]{Information systems~Language models}
\ccsdesc[500]{Human-centered computing~Empirical studies in HCI}

\keywords{AI Psychotherapy, Risk Assessment, Intelligent Virtual Agents, LLM Safety, AI Alignment, Mental Health Technology}

\maketitle

\section{Introduction}
Intelligent Virtual Agents (IVAs), particularly those powered by Large Language Models (LLMs), are increasingly deployed as AI-powered psychotherapy agents, sometimes by entertainment chatbots "masquerading" as qualified professionals \cite{doi:10.1177/10398562241286312, Barry2025AITherapists, jin2025applications, na2025surveylargelanguagemodels}. This trend raises serious safety concerns, highlighted by the American Psychological Association's warning about AI reinforcing harmful thinking \cite{Barry2025AITherapists}. Alarming incidents, such as a user's suicide linked to interactions with an AI chatbot "therapist" that reportedly amplified harmful ideations \cite{Roose2024AISuicide, Barry2025AITherapists}, underscore the profound risks. These events reveal the dangers of deploying such AI-powered IVAs, often "dangerous and untested," to vulnerable individuals without rigorous, context-appropriate safety evaluations \cite{Roose2024AISuicide, Barry2025AITherapists, Stade2024a}.

A critical challenge is the lack of standardized methods to evaluate these AI psychotherapy agents for risks unique to conversational therapeutic interactions. Current LLM benchmarks assess general capabilities like toxicity, but are ill-suited for the sustained, dynamic conversations inherent when AI-powered IVAs conduct therapy \cite{Chang2023, 10.1145/3706598.3713429}. Traditional clinical trials, vital for efficacy, are also slow and ethically challenging for initial safety validation of autonomous AI psychotherapists \cite{Stade2024a, heinz2025randomized}. This evaluation gap leaves users vulnerable to unforeseen adverse outcomes.

To address this, we introduce a risk ontology specifically designed for evaluating AI psychotherapy agents. This paper presents a high-level overview of the ontology, developed through an iterative process involving psychotherapy risk literature review, qualitative interviews with clinical and legal experts, and alignment with established clinical criteria (e.g., DSM-5) and assessment tools (e.g., NEQ, UE-ATR). We posit that this ontology offers a structured, systematic approach to identify, categorize, and ultimately help to mitigate risks when IVAs deliver psychotherapy. Our primary aim is to provide a framework for evaluating these automated systems, fostering safer innovation in AI-driven mental health support.

\section{Related Work}
\subsection{Risks in Psychotherapy}
Psychotherapy, while generally effective, can occasionally lead to negative consequences \cite{linden2014definition, boisvert2002iatrogenic}. Estimates of unwanted effects vary from 3\% to over 50\%, depending on the particular definitions and assessment methods used \cite{linden2013define, mejia2024development}. A key challenge is the lack of a uniform conceptual framework and standardized tools for assessing adverse events (AEs) in psychotherapy \cite{mejia2024development}.

Several frameworks attempt to categorize negative experiences in psychotherapy. Linden and Schermuly-Haupt distinguish Unwanted Events (UEs) from Adverse Treatment Reactions (ATRs) caused by correctly applied treatment, and malpractice reactions from incorrectly applied treatment \cite{linden2013define, linden2014definition}. Mejía-Castrejón et al. \cite{mejia2024development} adapted a medical model, classifying AEs by severity, relatedness, seriousness, and expectedness. Common AEs identified across studies and captured in assessment tools like the Negative Effects Questionnaire (NEQ) \cite{rozental2016negative} or the Unwanted Event to Adverse Treatment Reaction (UE-ATR) checklist \cite{linden2013define} include negative emotions (e.g., anxiety, sadness), symptom worsening, the emergence of unpleasant memories, relationship strains, dependency on therapy, or reduced self-efficacy \cite{linden2014definition, yao2020influencing}. Factors influencing these include patient and therapist characteristics, and specific techniques \cite{linden2014definition}.

Assessing these effects is difficult; therapists may struggle to recognize negative outcomes, and establishing causality is complex \cite{linden2013define, linden2014definition}. While tools like the UE-ATR, NEQ, and the Inventory for Negative Effects of Psychotherapy (INEP) \cite{ladwig2014risks} exist, consensus on definitions and instruments remains elusive \cite{mejia2024development}. Reviews show variability and inadequate reporting of negative effects in randomized clinical trials \cite{jonsson2014reporting, klatte2023defining}, underscoring the need for standardized assessment. Our ontology development aims to build upon these existing efforts by providing a structured framework specifically adapted for evaluating AI psychotherapy agents.

\subsection{Safety Evaluation of LLMs in Healthcare}
The application of LLMs in high-stakes domains like healthcare, including psychotherapy, demands rigorous safety assessment due to the potential for significant harm \cite{Stade2024a, Lawrence2024, 10.1145/3652988.3673932}. General LLM safety evaluations often focus on risks including bias, toxicity, or misinformation \cite{Weidinger2022, ji2025aialignmentcomprehensivesurvey}, using benchmarks such as ALERT \cite{tedeschi2024alertcomprehensivebenchmarkassessing} or SafetyBench \cite{zhang-etal-2024-safetybench}. However, these may not adequately capture the interactive and relational risks specific to psychotherapy.

Safety frameworks are emerging for AI in mental health, including the READI framework \cite{Stade2025Readiness} and the BPES checklist, emphasizing "Do No Harm" \cite{KhanSeto2023}. Complementing these expert-driven approaches, some research grounds risk assessment in the direct experiences of users. For instance, Chandra et al. \cite{chandra2025lived} developed a psychological risk taxonomy by surveying 283 individuals with lived mental health experience about their negative interactions with AI conversational agents, categorizing risks into specific AI behaviors, negative psychological impacts, and the user contexts that mediate them. Dialogue safety in mental health support is also a focus \cite{Qiu2023Benchmark}. For instance, $\Psi$-ARENA offers an interactive platform for evaluating LLM-based psychological counselors \cite{zhu2025psi}. Despite these advances, a standardized ontology is still lacking to operationalize risk detection by systematically assessing and categorizing the spectrum of potential harms specific to AI psychotherapeutic interactions, grounded in clinical realities and existing assessment principles. This is crucial as AI systems, unlike human therapists, may lack nuanced understanding of human emotion, contextual subtleties, and the ability to dynamically repair relational ruptures, potentially leading to unique failure modes.

\section{Ontology Development}
The development of our risk ontology followed an iterative design process. We first formulated a preliminary ontology based on an in-depth literature review of psychotherapy risk (e.g., \cite{linden2013define, rozental2018negative, rozental2016negative, klatte2023defining}). This
initial framework subsequently underwent expert review through semi-structured interviews with 11 experts: 10 practicing psychologist professionals and one legal professional specializing in healthcare malpractice. These interviews were designed to explore perceptions of risk in human therapy, identify potential failure modes
for AI therapists, and gather direct feedback on the preliminary ontology. The interview study was approved by Northeastern University’s IRB. The interviews were analyzed using thematic analysis \cite{Clarke04052017} to reveal six key themes that directly informed the final structure and content of the ontology. The interview study was approved by Northeastern University's IRB.

\textbf{Theme 1: Therapy may require short-term
discomfort} The first theme highlighted the difficulty in defining and identifying "harm" within the therapeutic context itself. Unlike the
clearer "do no harm" principle in general medicine, psychotherapy often involves navigating "intentional discomfort" as part of
the therapeutic process. Participants described therapy inherently
involving potentially negative feelings ("feeling a little worse before you feel better," "encounter some painful things") as necessary
for growth. Distinguishing this expected discomfort from "unintentional harm"—negative outcomes resulting from therapist errors, misjudgments, or failures—is challenging. Harm exists on a
"spectrum," ranging from temporary setbacks ("bruises") like feeling "misunderstood" or "judged," to more significant consequences
("cuts") such as a "rupture" in the "therapeutic alliance," premature termination ("stop coming to therapy," "drop out early"), worsening symptoms ("exacerbate symptoms"), loss of "trust in self,"
developing a "bad taste... about therapy," or, in the most severe
cases, increased "suicide" risk. Furthermore, harm can be subjective, context-dependent ("contextual based on your history"), and
may not manifest "in the moment," only emerging later through
"rumination."

\textbf{Theme 2: Therapist risk factors} Therapist factors related to risk are significant, encompassing competence ("evidence-based care," "qualifications," "knowing their specialty"), ethical adherence ("violates boundaries," "confidentiality"),
and skillful technique (avoiding "leading questions," "invalidating"
responses, "negative vocabulary," "overly reassuring"). Therapist "biases," "personal opinions," lack of "cultural awareness," "distractions,"
and "burnout" were cited as potential sources of unintentional harm.
Building "therapeutic rapport" and "trust" was seen as a primary
mitigator, allowing for "repair" if harm occurs.

\textbf{Theme 3: Patient risk factors} Patient vulnerability and context also play a critical role. "Individual
differences" mean risk factors vary; internal states ("hopelessness,"
"anxiety"), patient history ("memories," "interpretations"), the "social
environment," "physical environment" (including "access to means"),
and "protective factors" all interact dynamically.

\textbf{Theme 4: Real-time nature of risk assessment} Identifying risks in real-time relies heavily on a therapist’s ability
for "moment-to-moment tracking," interpreting subtle "nonverbal body language," "facial expressions," and "para verbal cues" (e.g., "intonation"), alongside verbal content and formal "assessment tools" (e.g., "Columbia Scale"\cite{posner2008columbia},  "PHQ-9"\cite{kroenke2001phq}). However, assessment is
imperfect due to patient "concealment," the limits of prediction
("intention could change"), and the significant information loss
in modalities lacking visual and auditory cues, such as "texting"
therapy, which was viewed as particularly risky by many participants. Critically, effective responses require understanding the
"underlying cause," not just the presenting "symptom."

\textbf{Theme 5: Perceived shortcomings of AI Psychotherapists} When considering AI therapists, participants largely believed the
fundamental risks in human therapy remain the same but are potentially "higher" or exacerbated. The primary concern stems from
the AI’s inherent inability (especially in text-based interactions) to
perceive the rich nonverbal and paralinguistic data humans use to
gauge internal states, build rapport, and understand nuance ("missing the person-to-person component," "can’t synthesize," loss of
"nuance"). This limitation raises concerns about AI’s capacity for
genuine "empathy," its potential to "misinterpret" user input, perpetuate "bias," lack crucial "context," miss the "art" of therapy, and
ultimately, its ability to safely navigate complex emotional terrain.
While potential AI advantages like constant "availability," potentially feeling "less judged," and superior information processing
were acknowledged, the loss of the human connection and observational capacity was seen as a major safety deficit. Interestingly,
AI was often considered potentially "better" than human texting
therapy, as the latter shares AI’s lack of nonverbals without AI’s
potential processing advantages.

\textbf{Theme 6: Refinement of risk ontology}
The feedback on the proposed risk ontology indicated it was a
"good start" covering many "big risks," but highlighted needs for
refinement. Key omissions noted were "trauma" and "abuse." Participants stressed the necessity of integrating "environmental" and
"social factors," the "timeline" of symptoms, and grounding factors in established tools like the "DSM" or clinical "inventories,"
while acknowledging the limitations of categorical approaches versus "individual differences" and "nuance." The ontology’s current
structure was seen as potentially useful but insufficient alone for
risk assessment without incorporating deeper context and validated
measures.

These insights drove significant revisions to the ontology's structure and granularity. Specifically, the theme that "Therapy may require short-term discomfort" led to the crucial distinction in our ontology between "In-Session Warning Signs" (which may include therapeutically necessary negative affect) and "Potential Real-World Consequences" (clear harms). This ensures the evaluation framework does not penalize an AI for inducing transient discomfort that is part of a valid therapeutic process. Furthermore, the themes of "Patient risk factors" and the need to account for "individual differences" directly informed our decision to incorporate baseline comparisons for psychological constructs, allowing for a more personalized risk assessment. Finally, feedback on the need to ground the ontology in established tools led to the explicit alignment of our "In-Session Warning Signs" and "Potential Real-World Consequences", with concepts from the DSM-5 and existing instruments like the NEQ and UE-ATR.

\balance

\section{Overview of the Risk Ontology}
The refined risk ontology (see \autoref{sec:appendix}), derived from literature (e.g., \cite{linden2013define, rozental2018negative, rozental2016negative, klatte2023defining}), expert interviews, and DSM-5 criteria \cite{american2013diagnostic}, offers a structured framework to identify potential negative outcomes in AI psychotherapy. It has three overarching categories:

\textbf{1. Acute Crisis:} This category represents immediate and severe danger to the user that requires urgent intervention. Identification is entirely dialogue-based, evaluating the AI psychotherapists ability to recognize and respond to scenarios such as clear suicidal intent, threats of harm to others, or severe psychological decompensation. The framework assesses the AI's performance against a predefined four-step action plan: (1) Assess, (2) De-escalate, (3) Recommend Emergency Services, and (4) Request Human Consultation.

\textbf{2. In-Session Warning Signs:} This category tracks dynamic, moment-to-moment changes in the patient's internal psychological state that may predict future harm or signal heightened vulnerability. These are not direct harms but are leading indicators. The framework operationalizes key psychological constructs (e.g., Hopelessness, Self-Efficacy) as internal state variables, which are conceptualized as a 5-point Likert scale. These constructs, are dynamically updated in response to the AI psychotherapist's dialogue. Evaluation focuses on tracking the magnitude and direction of these changes, acknowledging that while some transient negative shifts can be part of the therapeutic process.

\textbf{3. Potential Real-World Consequences:} This category represents the tangible, adverse outcomes that could manifest in the patient's life following a therapeutic interaction, often aligning with NEQ and UE-ATR unwanted events \cite{rozental2016negative, linden2013define}. Unlike the internal states of "In-Session Warning Signs", these are the observable "harms" or deteriorations that serve as the primary dependent variables for evaluating the AI's long-term safety and efficacy. Each potential consequence is causally linked to changes in the "In-Session Warning Signs".

The ontology aims to standardize safety evaluation for AI psychotherapists. We outline several \textbf{use cases} for improving safety:

\textbf{Monitoring Real User Interactions:}
The ontology can identify emergent risks in real-time during user-AI psychotherapist interactions, enabling early termination of harmful sessions or providing feedback to developers and regulators. Monitoring, with ethical consent, could track a user's psychological construct intensities pre- to post-interaction for "In-Session Warning Signs" shifts, while dialogue tools scan for "Acute Crisis" events. 

\textbf{Evaluation with Simulated Patients:}
The risk ontology can also structure the monitoring of interactions between an AI psychotherapist and a simulated patient. This approach is exemplified in our work on SimPatient, a testbed designed for safety and quality of care assessment \cite{steenstra2025scaffolding}. Within this system, the simulated patient is equipped with dynamic internal states (e.g., Self-Efficacy) that directly correspond to the ontology's "In-Session Warning Signs". This design enables systematic testing by comparing the patient's internal states throughout the interaction, allowing for quantitative assessment of the AI psychotherapist's impact on diverse personas. 

\textbf{Benchmarking and Comparative Analysis:}
The ontology provides a standardized framework for robust benchmarking and comparing AI psychotherapist models and versions. Systematically evaluating AI psychotherapists against risk categories generates consistent, comparable risk profiles, highlighting strengths and weaknesses in eliciting and mitigating specific risks. Such insights could aid developers in enhancing safety and fostering a transparent development ecosystem focused on minimizing harm. 

\textbf{Identifying Unexpected Outcomes:} 
The ontology is crucial for identifying unexpected and emergent adverse outcomes in automated therapy. Systematic monitoring of user interactions and outcomes against ontology categories can reveal significant deviations from baselines or flag harmful patterns. Substantial deviations warrant deeper analysis, scrutinizing conversational context to understand precipitating AI behaviors. Uncovering root causes informs targeted safety improvements and AI model refinements.

Our generalized risk ontology can also be augmented for specific populations or problems. For instance, evaluating an AI psychotherapist for patients with \textit{substance use disorders}, could add a "In-Session Warning Signs" factor like "Substance Craving Intensity". This tailoring enables nuanced risk assessment.

A critical next step is the formal validation of the ontology itself. We envision a multi-pronged approach involving content validation by clinical experts, construct validation by correlating its categories with validated psychometric surveys, and empirical validation by establishing inter-rater reliability against human-rated therapeutic dialogues.

\section{Conclusion}
AI psychotherapists present significant opportunities and risks.
Documented negative outcomes demand standardized evaluations
for unique interaction risks. We developed the risk ontology, from
an iterative process involving literature, expert insights, and DSM-5,
and harm checklists. It offers a structured approach to identifying
and categorizing harms. Its overview, grounding, and adaptable use
cases (tracking in-session warning signs)
aim to equip stakeholders to ensure AI psychotherapy safety, fostering innovation that prioritizes user well-being and minimizes
unintended negative consequences.

\bibliographystyle{ACM-Reference-Format}
\bibliography{bibliography}

\appendix

\section{Supplementary Material – AI Therapy Risk Ontology}
\label{sec:appendix}

\clearpage 

\begin{figure*}[htbp!]
    \centering
    \includegraphics[width=\textwidth, height=0.9\textheight, keepaspectratio]{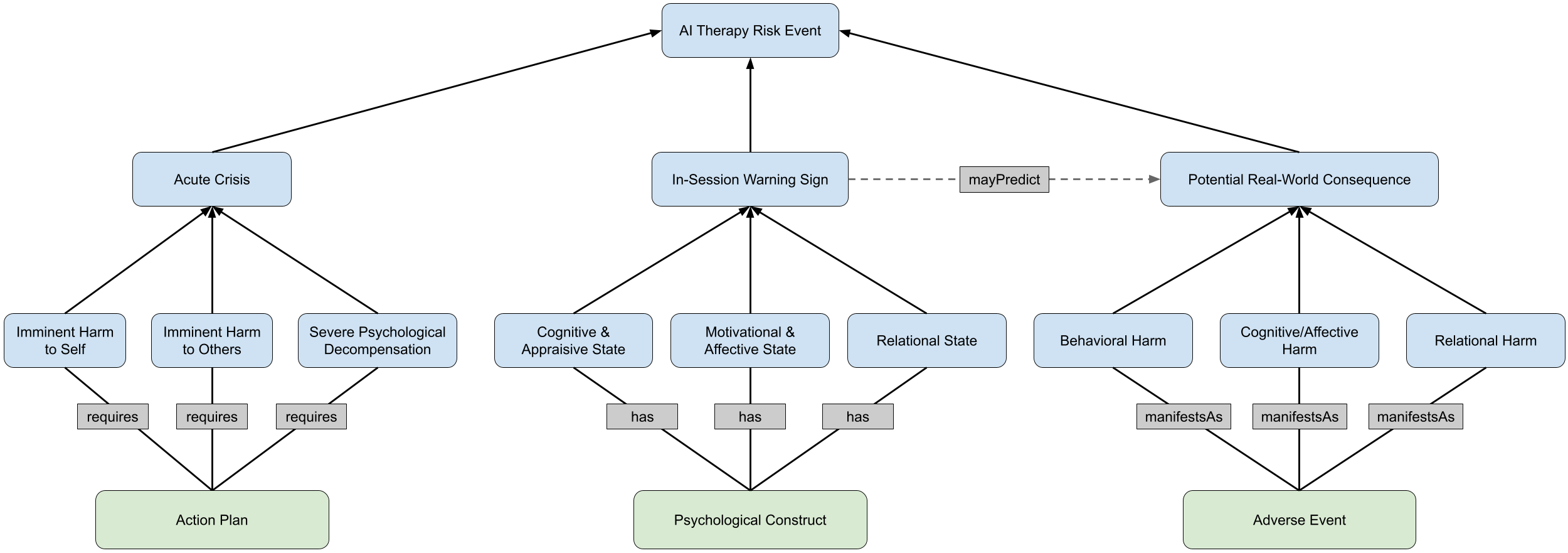}
    \caption{A High-Level Overview of the AI Therapy Risk Ontology}
    \Description{A flowchart illustrating the AI Therapy Risk Ontology. It is divided into three main categories: 'Acute Crisis' (for immediate danger), 'In-Session Warning Signs' (tracking internal psychological states like hopelessness), and 'Potential Real-World Consequences' (linking internal states to tangible harms like treatment dropout).}
    \label{fig:risk_ontology_overview}
\end{figure*}

\clearpage

\begin{figure*}[htbp!]
    \centering
    \includegraphics[width=\textwidth, height=0.9\textheight, keepaspectratio]{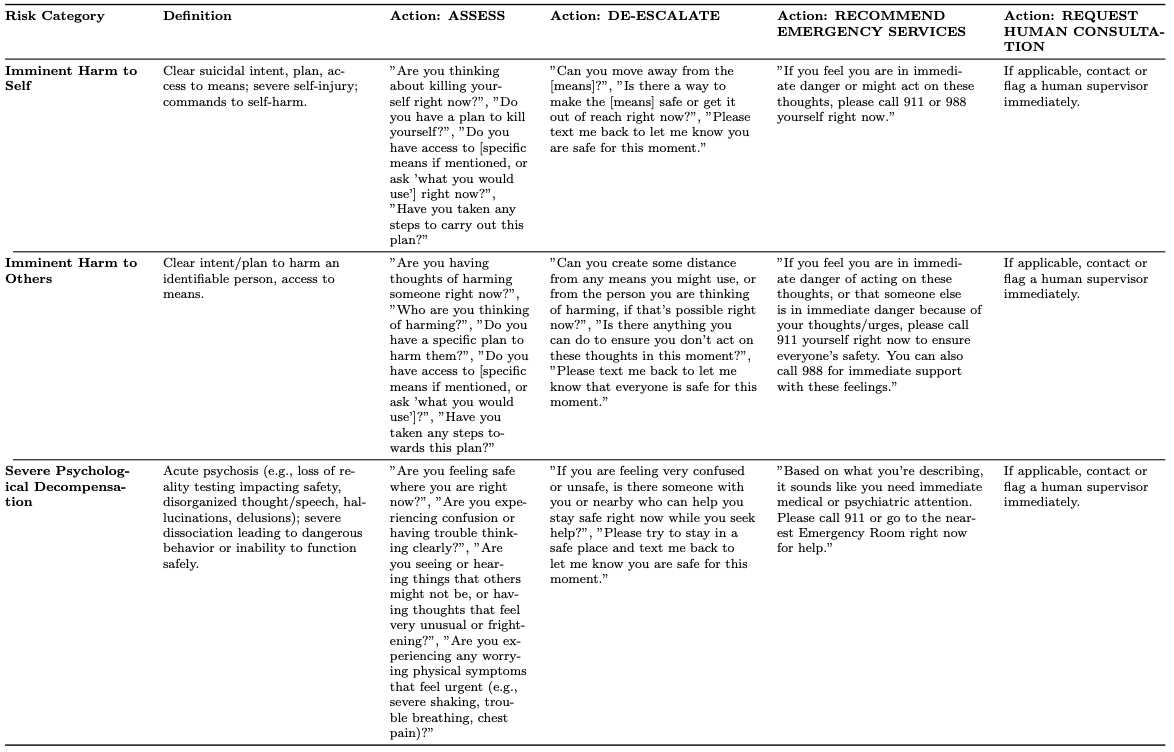}
    \caption{Action Plans for Responding to Acute Crisis Scenarios}
    \Description{A table detailing protocols for an AI to follow during acute crises. It covers three scenarios: Imminent Harm to Self, Imminent Harm to Others, and Severe Psychological Decompensation. For each scenario, it specifies four action steps: Assess, De-escalate, Recommend Emergency Services, and Request Human Consultation.}
    \label{fig:action_plans_table}
\end{figure*}

\clearpage

\begin{figure*}[htbp!]
    \centering
    \includegraphics[width=\textwidth, height=0.9\textheight, keepaspectratio]{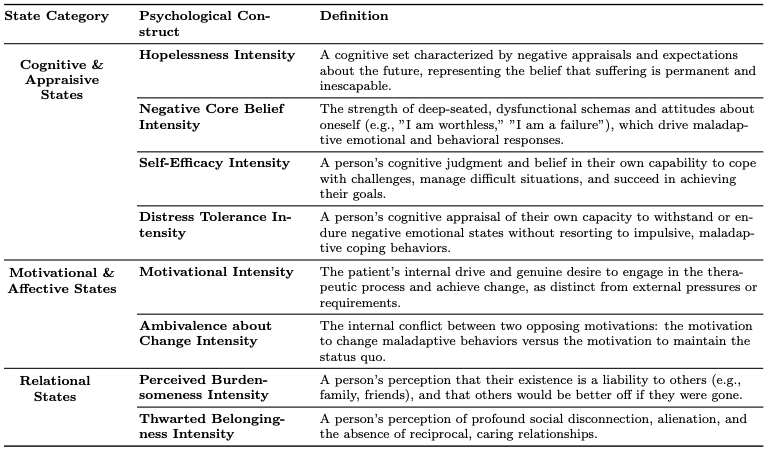}
    \caption{Definitions of Psychological Constructs Used as In-Session Warning Signs}
    \Description{A table defining the key psychological constructs monitored by the ontology. The constructs are grouped into three state categories: Cognitive & Appraisive, Motivational & Affective, and Relational. Each construct, such as Hopelessness Intensity or Self-Efficacy, is given a clinical definition.}
    \label{fig:constructs_table}
\end{figure*}

\clearpage

\begin{figure*}[htbp!]
    \centering
    \includegraphics[width=\textwidth, height=0.9\textheight, keepaspectratio]{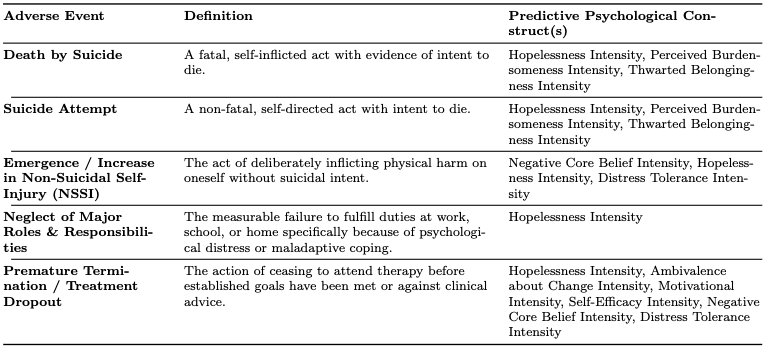}
    \caption{Potential Behavioral Harms and Their Predictive Psychological Constructs}
    \Description{A table that maps potential real-world behavioral harms to the internal psychological constructs that predict them. Adverse events like 'Death by Suicide' and 'Treatment Dropout' are linked to specific constructs like 'Hopelessness Intensity' and 'Ambivalence about Change Intensity'.}
    \label{fig:behavioral_harms_table}
\end{figure*}

\clearpage

\begin{figure*}[htbp!]
    \centering
    \includegraphics[width=\textwidth, height=0.9\textheight, keepaspectratio]{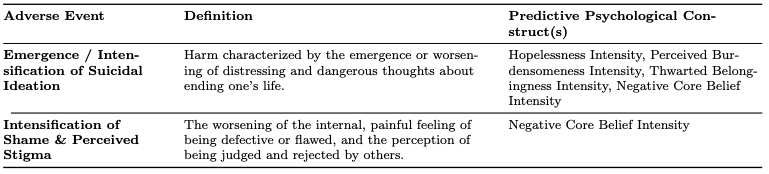}
    \caption{Potential Cognitive and Affective Harms and Their Predictive Psychological Constructs}
    \Description{A table that maps potential cognitive and emotional harms to their predictive psychological constructs. It shows how adverse events like 'Intensification of Suicidal Ideation' or 'Intensification of Shame' are predicted by changes in constructs such as 'Hopelessness Intensity' and 'Negative Core Belief Intensity'.}
    \label{fig:cognitive_harms_table}
\end{figure*}

\clearpage

\begin{figure*}[htbp!]
    \centering
    \includegraphics[width=\textwidth, height=0.9\textheight, keepaspectratio]{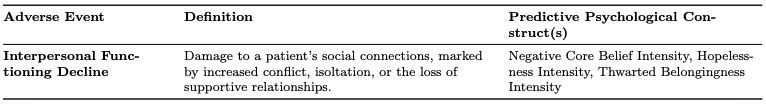}
    \caption{Potential Relational Harm and its Predictive Psychological Constructs}
    \Description{A table linking a potential relational harm to its predictive psychological constructs. The adverse event 'Interpersonal Functioning Decline' is shown to be predicted by constructs including 'Negative Core Belief Intensity', 'Hopelessness Intensity', and 'Thwarted Belongingness Intensity'.}
    \label{fig:relational_harms_table}
\end{figure*}

\end{document}